\documentclass[10pt,twocolumn,letterpaper]{article}

\usepackage{iccv}
\usepackage{times}
\usepackage{epsfig}
\usepackage{graphicx}
\usepackage{amsmath}
\usepackage{amssymb}
\usepackage{multirow}
\usepackage{caption}

\newcommand{\tablestyle}[2]{\setlength{\tabcolsep}{#1}\renewcommand{\arraystretch}{#2}\centering\footnotesize}
\newcommand{\cready}[1]{\textcolor{black}{#1}}
\newlength\savewidth\newcommand\shline{\noalign{\global\savewidth\arrayrulewidth
  \global\arrayrulewidth 1pt}\hline\noalign{\global\arrayrulewidth\savewidth}}
\newcommand{\myplus}[1]{\color{green}{\tiny{$+$#1}}}
\newcommand{\myminus}[1]{\color{red}{\tiny{$-$#1}}}

\usepackage[pagebackref=true,breaklinks=true,letterpaper=true,colorlinks,bookmarks=false]{hyperref}

\iccvfinalcopy 

\def\iccvPaperID{3170} 

\ificcvfinal\fi

\begin{document}

\title{Exploring Inter-Channel Correlation for Diversity-preserved Knowledge Distillation}

\author{Li Liu$^{1}$\footnotemark[2]\; Qingle Huang$^{1}$\footnotemark[2] \; Sihao Lin$^{2}$\footnotemark[2]\, \footnotemark[3] 
\; Hongwei Xie$^1$ \; Bing Wang$^1$ \; Xiaojun Chang$^3$\thanks{Corresponding Author.} 
\; Xiaodan Liang$^4$ \\
$^1$Alibaba Group\;
$^2$Monash University\;
$^3$RMIT University\;
$^4$Sun Yat-sen University\\

{\tt\small  \{liuli.ll9412, qingle.hql, linsihao6, hongwei.xie.90, xdliang328\}@gmail.com}\\
{\tt\small fengquan.wb@alibaba-inc.com, xiaojun.chang@rmit.edu.au}

}

\maketitle
\ificcvfinal\thispagestyle{empty}\fi

\begin{abstract}
    Knowledge Distillation has shown very promising ability in transferring learned representation from the larger model (teacher) to the smaller one (student). Despite many efforts, prior methods ignore the important role of retaining inter-channel correlation of features, leading to the lack of capturing intrinsic distribution of the feature space and sufficient diversity properties of features in the teacher network. To solve the issue, we propose the novel Inter-Channel Correlation for Knowledge Distillation (ICKD), with which the diversity and homology of the feature space of the student network can align with that of the teacher network. The correlation between these two channels is interpreted as diversity if they are irrelevant to each other, otherwise homology. Then the student is required to mimic the correlation within its own embedding space. In addition, we introduce the grid-level inter-channel correlation, making it capable of dense prediction tasks. Extensive experiments on two vision tasks, including ImageNet classification and Pascal VOC segmentation,  demonstrate the superiority of our ICKD, which consistently outperforms many existing methods, advancing the state-of-the-art in the fields of Knowledge Distillation. To our knowledge, we are the first method based on knowledge distillation boosts ResNet18 beyond 72\% Top-1 accuracy on ImageNet classification. Code is available at: {\tt \small https://github.com/ADLab-AutoDrive/ICKD}.
\end{abstract}
\footnotetext[2]{Equal contribution.}
\footnotetext[3]{Work done when as an intern in DAMO Academy, Alibaba Group.}
\vspace{-6mm}
\section{Introduction}

It is widely witnessed that larger networks are superior in learning capacity compared to smaller ones. Nevertheless, due to the great amount of energy consumption and computation costs, a large network (\eg, ResNet-50 \cite{He2016DeepRL}), though powerful, is difficult to deploy on mobile systems. Hence, there is a growing interest in reducing the model size while preserving comparable performance, which bridges the gap between small networks and large networks.
\begin{figure}[t]
\begin{center}
  \includegraphics[width=0.8\linewidth]{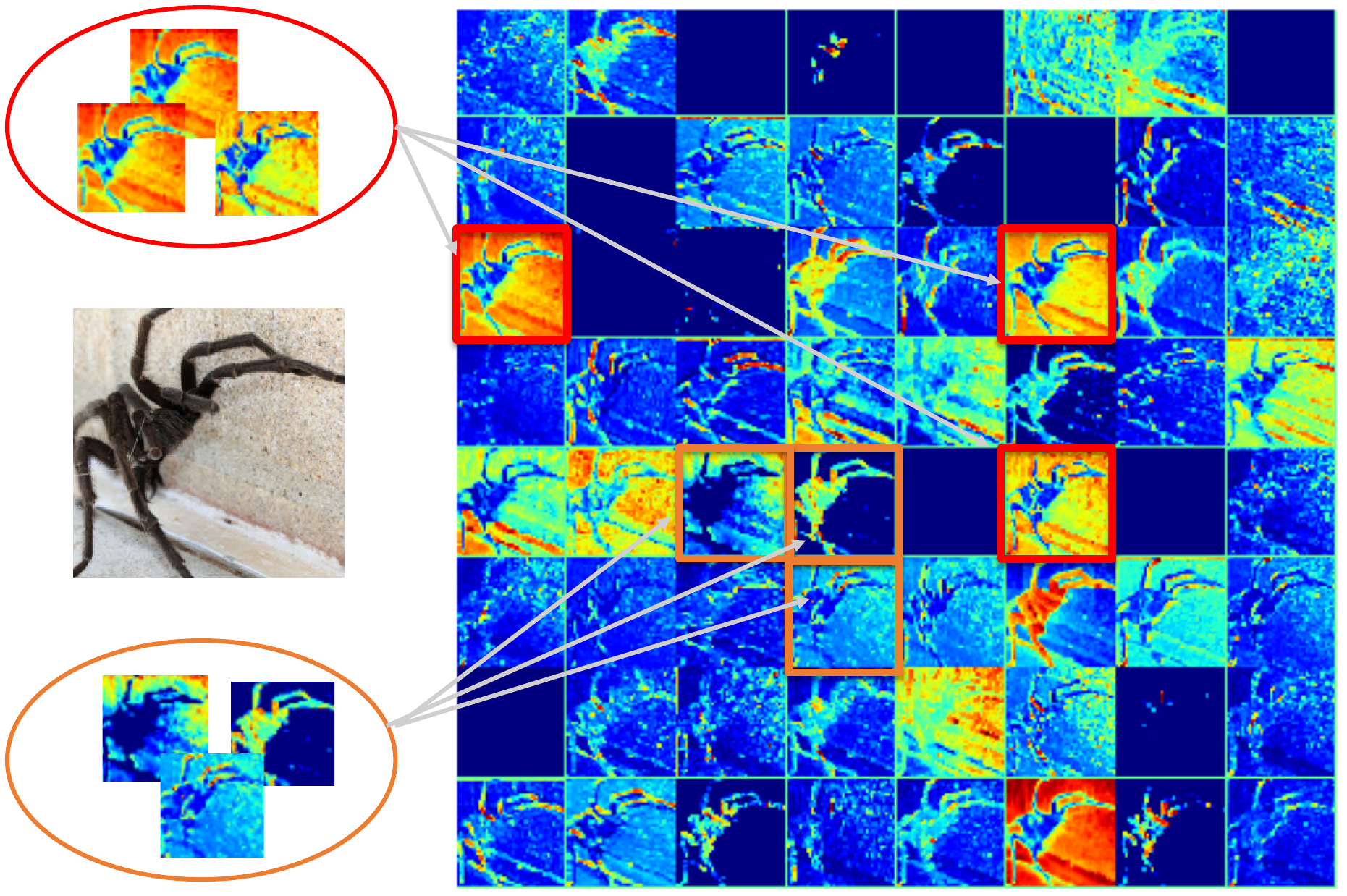}
\end{center}
  \vspace{-5mm}
  \caption{\textbf{Illustration of inter-channel correlation}. The channels orderly extracted from the second layer of ResNet18 have been visualized. The channels denoted by red boxes are homologous both perceptually and mathematically (\eg, inner-product), while the channels denoted by orange boxes are diverse. We show the inter-channel correlation can effectively measure that each channel is homologous or diverse to others, which further reflects the \textit{richness} of the feature spaces. Based on this insightful finding, our ICKD can enforce the student to \textit{mimic} this property from the teacher.}
\label{intro}
\vspace{-5mm}
\end{figure}

Knowledge distillation is one of the promising methods to this problem. 
It is acknowledged that Bucila \etal \cite{bucilua2006model} introduced the 
idea of knowledge distillation and Hinton \etal \cite{Hinton2015DistillingTK} further popularized 
this concept. The key idea of knowledge distillation is to let the student network \textit{mimic} 
the teacher model. The underlying principle is that teachers can provide the 
knowledge that ground truth labels can not tell. Despite its success, this 
technique, devoted to instance-level classification, may lead the student to mainly learn the instance-level information but not structural information, which limits its application. 
Prior works \cite{Park2019RelationalKD,Passalis2018LearningDR,Peng2019CorrelationCF,Tian2020ContrastiveRD,Tung2019SimilarityPreservingKD}
have been proposed to help the student network learn the structural representation for better generalization ability. 
These methods generally utilize the correlation of the instances to describe the geometry, similarity, or dissimilarity in the feature space. 
We call this fashion layer-wise relational knowledge distillation since they mainly focus on exploring the correlation between feature maps in the level of layer. Conversely, we pay more attention to the inter-channel correlation.

\begin{figure*}[!t]
\begin{center}
\includegraphics[width=0.8\linewidth]{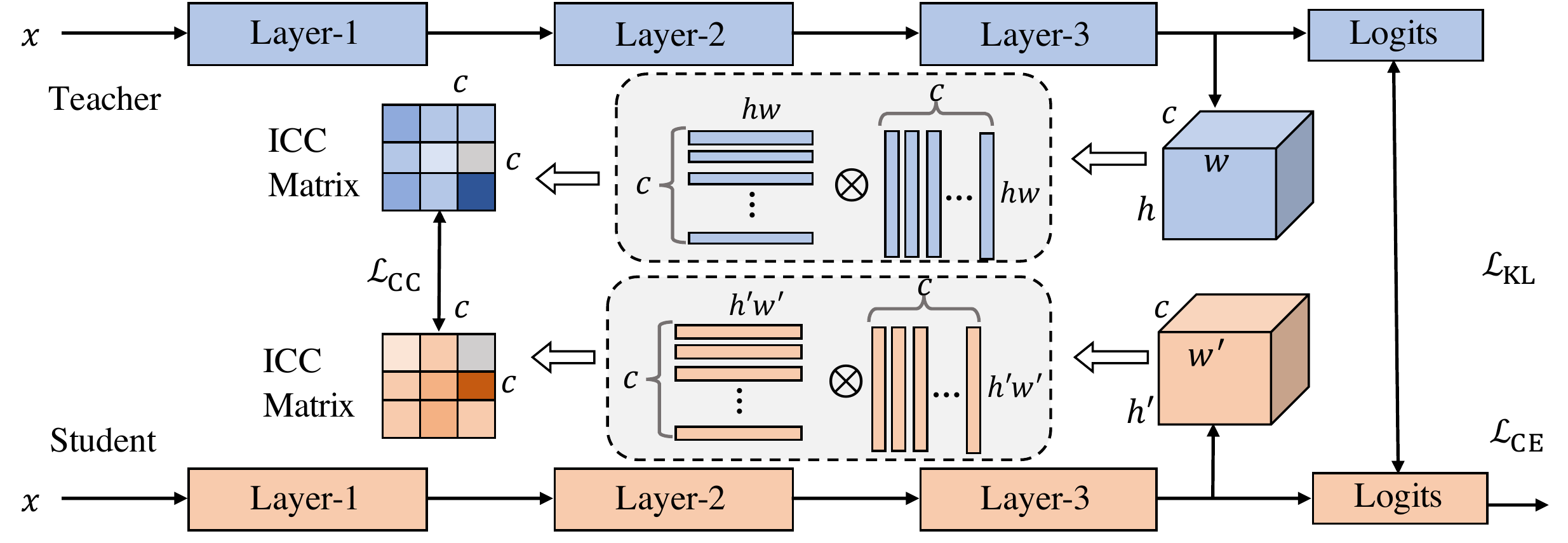}
\vspace*{-3mm}
\end{center}
   \caption{\textbf{Illustration of the proposed ICKD}. We measure the inter-channel correlation of the teacher feature and ask the student to share with the same property. The cubes represent the 3D feature tensors extracted from the teacher and student. They are flattened to the corresponding 2D matrices which are used to compute the ICC matrices. We minimize the MSE between the ICC matrices associated with the feature tensors. The student is also asked to minimize the KL-divergence between the logits of the teacher and student. Finally the cross-entropy loss is applied on the student.}
\label{fig:illustration}
\vspace{-4mm}
\end{figure*}

Previous works \cite{lee2018self,Sun2019PatientKD} make use of knowledge distillation to reduce the homology (\ie, redundancy) of the student's feature space. 
Nevertheless, the success of GhostNet \cite{Han2020GhostNetMF} suggests that small neural networks benefit from increased feature homology.
The rich representation can empower the downstream tasks and both the diversity and homology can reflect the richness.
Existing literature neglects the importance of feature diversity and homology, yielding an issue that the proportion of feature diversity versus homology may be unbalanced against our expectation that student can learn the representation as rich as the teacher is for better generalization. 
In Fig. \ref{intro}, the visualized feature maps show that feature diversity and homology co-exist in the networks. 
This property can be disclosed by the correlation between channels, where high relevance represents homology and low relevance represents diversity.
In this paper, we adopt the Inter-Channel Correlation (\textbf{ICC}) as the indicator of the diversity and homology of the feature distribution.
However, figuring out the optimal inter-channel correlation manually is impractical. 
An intuitive solution is to let the student learn better inter-channel correlation from the teacher, as shown in Fig. \ref{fig:illustration}.
Due to the discrepancy of learning capacities \cite{Cho2019OnTE}, it is not viable to force the student to mimic the whole feature map of the teacher. Instead, we let the student model learn the inter-channel correlation from the teacher, namely inter-channel correlation knowledge distillation (\textbf{ICKD}).

The correlation between the two channels is evaluated by the inner product in this paper. 
As the inner product collapses the spatial dimension, it naturally does not need to constrain the feature map spatial size of the teacher network and student network to be the same.
On the other hand, when it comes to a large feature map, \eg semantic segmentation models, the mapping between the inter-channel correlation measured by the inner product and the original feature space is of high freedom. Thus it will be more difficult to distill the inter-channel correlation distribution to anchor the teacher's feature space distribution.
To alleviate this problem, we propose a grid-level inter-channel correlation distillation method. By dividing the feature map of size  $h \times w \times c$ by a pre-defined grid into $n \times m$ patches of size $h_G \times w_G \times c$. 
Distillation on patch-level is more controllable and we can perform the distillation on the entire feature map by aggregating the inter-channel correlation distillation across these patches.
In addition, the local spatial information can be preserved since each patch can keep the knowledge in specific region.

 In our experiments, we have evaluated our proposed method in different tasks including classification (Cifar-100 and ImageNet) and semantic segmentation (Pascal VOC). 
 The proposed method shows a performance superior to the existing state-of-the-arts methods. To our knowledge, we are the first knowledge distillation method that boosts ResNet18 beyond 72\% Top-1 accuracy on ImageNet classification. And on Pascal VOC, we achieved 3\% mIoU improvement compared to the baseline model.

To summarize, our contributions are:
\vspace{-2mm}
\begin{itemize}
    \item  We introduce the inter-channel correlation, with the characteristic of being invariant to the spatial dimension, to explore and measure both the feature diversity and homology to help the student for better representation learning.
    \vspace{-2mm}
    \item We further introduce the grid-level inter-channel correlation to make our framework capable of dense prediction task, like semantic segmentation.
    \vspace{-2mm}
    \item To validate the effectiveness of the proposed framework, extensive experiments have been conducted on different (a) network architectures, (b) downstream tasks and (c) datasets. Our method consistently outperforms the state-of-the-arts methods by a large margin across a wide range of knowledge transfer tasks. 
    \vspace{-1mm}
\end{itemize}
\section{Related Works}\label{Related Works}
\noindent \textbf{Knowledge Distillation.} Given by \cite{Hinton2015DistillingTK}, the student network is required to minimize the KL-divergence between the logits (before softmax) output by the student and teacher, where a temperature $\tau$ is applied to soften the logits. This procedure, making it different from the ground truth
label, will increase the low probability in the logits, which is referred to \textit{dark knowledge}. 

To learn more generic representation, recent works \cite{Park2019RelationalKD,Passalis2018LearningDR,Peng2019CorrelationCF,Tian2020ContrastiveRD,Tung2019SimilarityPreservingKD} explored the structural information within the feature space of the teacher and transferred it to the student. 
Tung and Mori \cite{Tung2019SimilarityPreservingKD} measured the similarity among the given instances in the teacher's feature space and asked the student to match the same similarity. Peng \etal \cite{Peng2019CorrelationCF} presented the kernel-based correlation congruence. A kernel function was employed to measure the correlation metric of each paired instances in the feature space. Similarly, the student is required to share the same correlation metric with the teacher. RKD \cite{Park2019RelationalKD} further introduced the angle-wise relation given a triplet of instances. More recently, Tian \etal \cite{Tian2020ContrastiveRD} introduced contrastive learning to maximized the mutual information between the representation of the student and teacher.

Romero \etal \cite{Romero2015FitNetsHF} proposed the distillation in the
intermediate layers between the student (\ie, \textit{guided} layer) and
the teacher (\ie, \textit{hints} layer). The student is taught to
minimize the Euclidean distance of the feature maps from guided layer and
hints layer. 
Because the semantic information contained within the feature varies from layer to layer according to depth and width, existing works \cite{Ji2021ShowAA,Zagoruyko2017PayingMA,Chen2020CrossLayerDW}
has shown that layer-wisely match a pair
of guided layer and hints layer may not be an optimal choice. 
AT \cite{Zagoruyko2017PayingMA} proposed a statistical method to highlight the attention, compressing the 3D feature tensor to a 2D feature map. Chen \etal \cite{Chen2020CrossLayerDW} proposed semantic calibration to assign the target teacher layer to the student layer across layers depending on the inner-products of the teacher layers and the student layers. 
Ji \etal \cite{Ji2021ShowAA} measured the similarities, bounded to 1 with a softmax function, between the teacher and student features, which was used as the weights to balance the feature matching.

\noindent \textbf{Semantic Segmentation.} 
In spite of great challenge, some approaches based on knowledge distillation had been proposed in semantic segmentation. He \etal \cite{he2019knowledge} pre-trained an auto-encoder to match the features between the student and teacher, which also measured the affinity matrix of the paired instances in the teacher network and transferred it to the student network. Liu \etal \cite{liu2019structured} proposed the structured knowledge distillation consisting of the pair-wise similarity transfer and pixel-wise distillation like \cite{Hinton2015DistillingTK}. Liu \etal also transferred the holistic knowledge via adversarial learning. Wang \etal \cite{Wang2020IntraclassFV} proposed the Intra-class Feature Variation Distillation that also measured the pair-wise similarity between the features of each pixel and that of the corresponding class-wise prototype. Heo \etal \cite{Heo2019ACO} proposed a distillation loss with a designed margin ReLU to boost the performance of a student on semantic segmentation.

We further extend the framework with the grid-level inter-channel correlation for stabilizing the distillation process and preserving the spatial information.
Perhaps our work is most close to Huang and Wang \cite{huang2017like} which utilizes the Gram Matrix \cite{Gatys2016ImageST}. Yim \etal \cite{Yim2017AGF} proposed the flow of solution procedure that computed the Gram matrix across layers. The difference is that \cite{huang2017like,Yim2017AGF} measure the relation between pixel-wise positions and we explore the correlation between two channels. 
\section{Method}
In this section, we first briefly introduce the preliminary of knowledge distillation. Then we formulate the proposed method to show how we can compute the ICC matrix. Finally, we extend the framework with the grid-level inter-channel correlation.
\subsection{Preliminary}

Let $\mathcal{X}^N$ denote a set of distinct examples with cardinality $N$. Suppose that we have a teacher model $T$ and a student model $S$, which are denoted by $f^T$ and $f^S$, respectively. In practice, $f^T$ and $f^S$ can be any differential function and we parameterize them as convolutional neural network (CNN) here. $F^T\in \mathbb{R}^{c\times h\times w}$ represents the embedding in the teacher network, where $c$ is the number of output channels, $h$ and $w$ represents the height and width of the feature map. Similarly, let ${F^S}\in \mathbb{R}^{{c'}\times {h'}\times{w'}}$ denote the embedding in the student network. In general, traditional knowledge distillation attempts to minimize the divergence between the embedding of the student and the teacher,  in \cite{Hinton2015DistillingTK} the formulation can be described as:
\begin{equation} \label{eq:kl}
    {\mathcal{L}}_{\rm KL}=\frac{1}{N}\sum_{i=1}^N{D_{\rm KL}(\sigma(\frac{f^T(x_i)}{\tau}),\sigma(\frac{f^S(x_i)}{\tau}))},    
\end{equation}
\noindent where $D_{\rm KL}(\cdot,\cdot)$ measures the Kullback-Leibler divergence, $\sigma(\cdot)$ is the softmax function, $\tau$ is the temperature factor, $f^T(x)$ and $f^S(x)$ represent the outputs of the penultimate layer (before softmax) in the teacher network and the student network, respectively.

\subsection{Formulation}
\label{sec:formulation}
In this section we introduce the formulation of inter-channel correlation. Given two channels, the correlation metric should return a value reflecting their relevance. A high value indicates homologous otherwise diverse. Ultimately all the correlation metrics are gathered sequentially to represent the holistic diversity of the channels. The inter-channel correlation can be defined by
\vspace{-2mm}
\begin{equation}
    \mathcal{G}^{F^T}_{m,n}=K({\rm v}(F^T_m),{\rm v}(F^T_n)),
    \label{eq:eq2}
    \vspace{-2mm}
\end{equation}
where $F^T_m\in\mathbb{R}^{h\times w}$ denotes the $m$-th channel of the feature $F^T$, ${\rm v}(\cdot)$ vectorizes a 2D feature map into a vector with length $hw$, and $K(\cdot)$ is a function that measures the correlation of a input pair, where inner product is employed. Note that Eq. \ref{eq:eq2} returns a scalar in spite of the spatial dimensions of the channel. This can be rewritten in a manner of matrix multiplication, forming our ICC matrix:
\vspace{-2mm}
\begin{equation}
    \mathcal{G}^{F^T}={\rm f}(F^T)\cdot{\rm f}(F^T)^\top,
    \label{eq:eq3}
    \vspace{-2mm}
\end{equation}
\noindent where ${\rm f}(F^T) \in \mathbb{R}^{c\times hw}$ flattens the spatial dimensions. The resulting ICC matrix has a size of $c\times c$ regardless of the spatial dimensions $h$ and $w$. Following the empirical setting \cite{wang2019distilling,he2019knowledge}, we add a linear transformation layer $C_l$ on top of the feature of the student, which consists of a convolution layer with $1\times1$ kernels and a BN layer without activation function. In case that the output dimension $c'$ of the student mismatches with that of the teacher, $C_l$ can adapt $F^S$ to match the output dimension $c$ of $F^T$. This procedure would not change the spatial dimensions.
We penalize the L2 distance between the ICC matrices of the student and the teacher, allowing the student to obtain similar feature diversity.
\vspace{-3mm}
\begin{equation}
\begin{aligned}
    \mathcal{L}_{\rm CC}&=\frac{1}{c^2}||\mathcal{G}^{{C_l}(F^S)}-\mathcal{G}^{F^T}||_2^2 .
\end{aligned}
\label{eq:L_CC}
\vspace{-1mm}
\end{equation}

We refer the method described above as \textit{\textbf{ICKD-C}}, which is mainly developed for image classification. Finally, the objective of our method is given by
\vspace{-2mm}
\begin{equation}\label{eq:objective}
\begin{aligned}
    \mathcal{L}_{\rm ICKD-C}&=\mathcal{L}_{\rm CE}+\beta_1\mathcal{L}_{\rm KL}+\beta_2\mathcal{L}_{\rm CC},
\end{aligned}
\vspace{-2mm}
\end{equation}
where $\mathcal{L}_{\rm CE}$ is the cross-entropy loss, $\beta_1$ and $\beta_2$ are the weight factors.

\begin{figure}[t]
\begin{center}
  \includegraphics[width=0.8\linewidth]{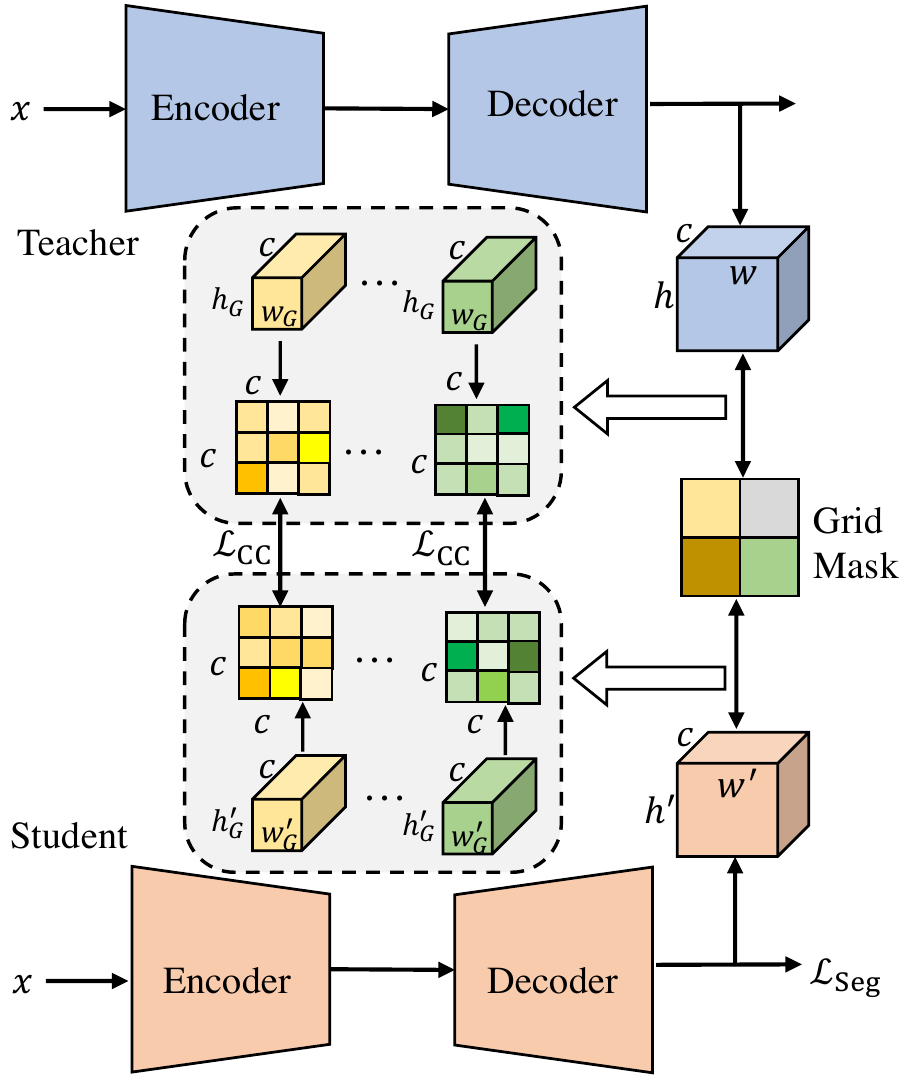}\vspace{-3mm}
\end{center}
  \caption{\textbf{Grid-level Inter-Channel Correlation}. We evenly divide the original feature into $n\times m$ parts and compute their ICC matrices individually. We then minimize the MSE on each paired ICC matrices.}
\label{fig:spatial_sampling}
\vspace{-4mm}
\end{figure}

\subsection{Grid-Level Inter-Channel Correlation}\label{sec:seg}
In Eq. \ref{eq:eq3}, we simply flatten the entire 3D feature map into the corresponding 2D matrix and then calculate the ICC matrix. 
When coming into semantic segmentation, the final feature map can be very large, \eg 256$\times$128$\times$128.
The correlation of two channels is generated by the inner-product of two vectors of length 16,384.
Generally, these two vectors can be seen as sampled from an independent distribution, the correlation value will be of a very small order of magnitude, that means the correlation result is vulnerable to noise.
In this situation, the inter-channel correlation of the student model in training process may be unstable.
Motivated by the divide-and-conquer, we seek to split the feature map and then perform knowledge distillation individually.

\begin{table*}[htb]
\centering
\caption{Top-1 accuracy (\%) in Cifar-100 testing set. Methods are divided into two groups. The performance of each method against traditional KD \cite{Hinton2015DistillingTK} is reported. For fair comparison, we also report the performance of our method without  $\mathcal{L}_{\rm KL}$. We find that our ICKD-C outperforms all the other methods.}
\resizebox{2.0\columnwidth}{!}{\tablestyle{10pt}{1.05}
\begin{tabular}{@{}l|ccccccc@{}} \shline
\multirow{3}{*}{Method} & \multicolumn{7}{c}{Network Architecture}  \\ \cline{2-8}
& WRN-40-2 & WRN-40-2  & ResNet56 & ResNet110 & ResNet110 & ResNet32$\times$4 & VGG13 \\
& WRN-16-2 & WRN-40-1  & ResNet20 & ResNet20  & ResNet32  & ResNet8$\times$4  & VGG8 \\ \hline
Teacher &75.61\hspace{5mm} &75.61\hspace{5mm} &72.34\hspace{5mm} &74.31\hspace{5mm} &74.31\hspace{5mm} &79.42\hspace{5mm} &74.64\hspace{5mm} \\ 
Vanilla &73.26\hspace{5mm} &71.98\hspace{5mm} &69.06\hspace{5mm} &69.06\hspace{5mm}	&71.14\hspace{5mm} &72.50\hspace{5mm} &70.36\hspace{5mm} \\  
KD \cite{Hinton2015DistillingTK}&74.92\hspace{5mm} &73.54\hspace{5mm} &70.66\hspace{5mm} &70.67\hspace{5mm} &73.08\hspace{5mm} &73.33\hspace{5mm} &72.98\hspace{5mm} \\ 
FitNet \cite{Romero2015FitNetsHF}&73.58\myminus{1.34}&72.24\myminus{1.30}&69.21\myminus{1.45}&68.99\myminus{1.68}&71.06\myminus{2.02}&73.50\myplus{0.17}&71.02\myminus{1.96} \\ 
AT \cite{Zagoruyko2017PayingMA}&74.08\myminus{0.84}&72.77\myminus{0.77}&70.55\myminus{0.11}&70.22\myminus{0.45}&72.31\myminus{0.77}&73.44\myplus{0.11}&71.43\myminus{1.55} \\ 
SP \cite{Tung2019SimilarityPreservingKD}&73.83\myminus{1.09}&72.43\myminus{1.11}&69.67\myminus{0.99}&70.04\myminus{0.63}&72.69\myminus{0.39}&72.94\myminus{0.39}&72.68\myminus{0.20} \\ 
CC \cite{Peng2019CorrelationCF}&73.56\myminus{1.36}&72.21\myminus{1.33}&69.63\myminus{1.03}&69.48\myminus{1.19}&71.48\myminus{1.6}&72.97\myminus{0.36}&70.71\myminus{2.27} \\ 
RKD \cite{Park2019RelationalKD}&73.35\myminus{1.57}&72.22\myminus{1.32}&69.61\myminus{1.05}&69.25\myminus{1.42}&71.82\myminus{1.26}&71.90\myminus{1.43}&71.48\myminus{1.5} \\
PKT \cite{Passalis2018LearningDR}&74.54\myminus{0.38}&73.45\myminus{0.09}&70.34\myminus{0.32}&70.25\myminus{0.42}&72.61\myminus{0.47}&73.64\myplus{0.31}&72.88\myminus{0.10} \\ 
FSP \cite{Yim2017AGF}&72.91\myminus{2.01}&NA&69.95\myminus{0.71}&70.11\myminus{0.56}&71.89\myminus{1.19}&72.62\myminus{0.71}&70.20\myminus{2.78}\\
NST \cite{huang2017like}&73.68\myminus{1.24}&72.24\myminus{1.3}&69.60\myminus{1.06}&69.53\myminus{1.14}&71.96\myminus{1.12}&73.30\myminus{0.03}&71.53\myminus{1.45}\\ \hline
ICKD-C (w/o $\mathcal{L}_{\rm KL}$)&\textbf{75.64}\myplus{0.72}&74.33\myplus{0.79}&\textbf{71.76}\myplus{1.1}&71.68\myplus{1.01}&73.89\myplus{0.81}&75.25\myplus{1.92}&73.42\myplus{0.44} \\
ICKD-C (Ours)&75.57\myplus{0.65}&\textbf{74.63}\myplus{1.09}&71.69\myplus{1.03}&\textbf{71.91}\myplus{1.24}&\textbf{74.11}\myplus{1.03}&\textbf{75.48}\myplus{2.15}&\textbf{73.88}\myplus{0.9}\\
 \hline
\end{tabular}
}\label{tab:cifar_sota}
\vspace{-4mm}
\end{table*}

Based on this idea, we introduce the grid-level inter-channel correlation. We evenly partition the feature $F^T$ into $n\times m$ parts along the pixel position, denoted by $F^T_{(i,j)}$, $i=1,2,...,n$, $j=1,2,...,m$. Each part is of size $c\times h_G \times w_G$ where $h_G=h/n$ and $w_G=w/m$. Each part presents the semantic on a patch level. The ICC matrix of each part is computed individually as described in Sec. \ref{sec:formulation}. Then all the ICC matrices are aggregated. 
\begin{equation}
    \mathcal{G}^{F^T_{(i,j)}}={\rm f}(F^T_{(i,j)})\cdot{\rm f}(F^T_{(i,j)})^\top,
\end{equation}
\begin{equation}
    \mathcal{G}^{F^S_{(i,j)}}={\rm f}(F^S_{(i,j)})\cdot{\rm f}(F^S_{(i,j)})^\top,
\end{equation}
\begin{equation}
     \mathcal{L}^{n\times m}_{\rm CC}=\frac{1}{{n \times m} \times {c^2}}\sum^n_i \sum^m_j ||\mathcal{G}^{F^T_{(i,j)}}-\mathcal{G}^{F^S_{(i,j)}}||_2^2.
    \label{eq:seg_cc}
\end{equation}

As depicted in Fig. \ref{fig:spatial_sampling}, we use a Grid Mask to evenly divide the whole feature into different groups. Despite the change of spatial dimensions, the size of the resulting ICC matrix always depend on the numbers of channels, \ie, $c$. 
In addition, the grid division also helps to extract more spatial and local information, which is beneficial in correctly classifying pixels for semantic segmentation \cite{wei2019building}.
This variant is referred to \textbf{\textit{ICKD-S}}. Finally, the objective for semantic segmentation is formulated as:
\begin{equation}\label{eq:seg_objective}
\begin{aligned}
    \mathcal{L}_{\rm ICKD-S}&=\mathcal{L}_{\rm Seg}+\alpha\mathcal{L}^{n\times m}_{\rm CC},
\end{aligned}
\end{equation}
\noindent where $\alpha$ is the weight factor and $\mathcal{L}_{\rm Seg}$ is the supervised segmentation loss. $\mathcal{L}_{\rm Seg}$, though, can be replaced with other loss for different downstream tasks, this is not the focus of this paper.

\begin{figure*}[!th]
    \centering
    \includegraphics[width=1.0\linewidth]{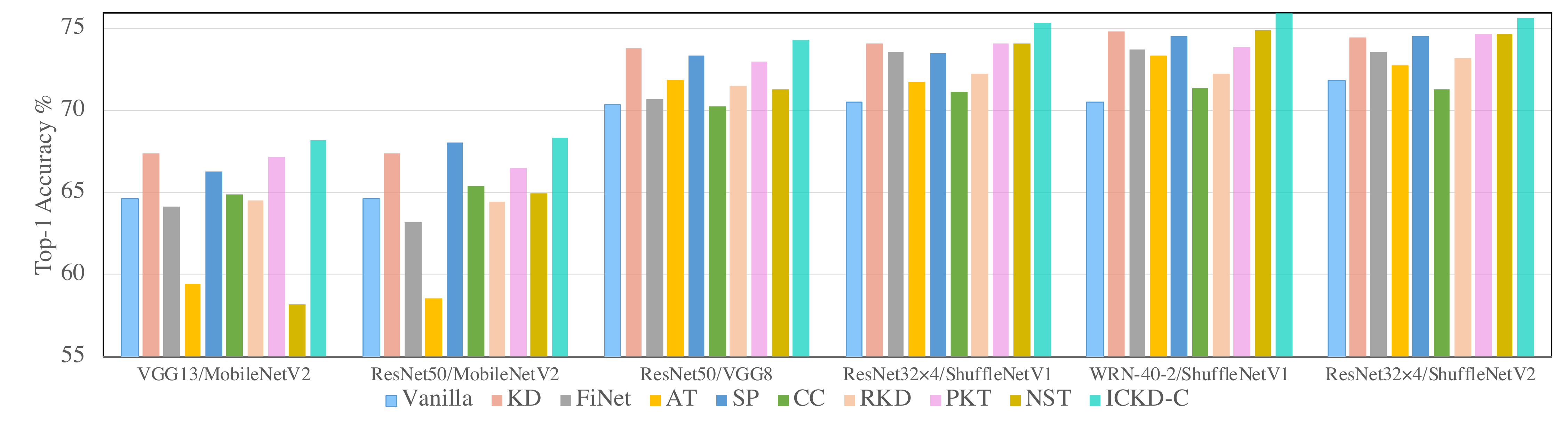}
    \vspace{-8mm}
    \caption{\textbf{Knowledge distillation across different architectures on Cifar-100.} Using teacher networks that completely different from that of students for knowledge distillation. \cready{The model before the slash is the teacher and the one after is the student.} Our method can enable the students to learn more general knowledge regardless of the specific architecture.}
    \label{fig:diff_arch}
\end{figure*}

\begin{table*}[htb]
\centering
\caption{Top-1 and Top-5 Accuracy (\%) on ImageNet validation set. The teacher network is ResNet34 and the student network is ResNet18. Our method outperforms other state-of-the-arts by a significant margin. Methods denoted by * do not release Top-5 accuracy.}
\vspace{-3mm}
\resizebox{2.0\columnwidth}{!}{
\begin{tabular}{@{}l|cccccccccc|c@{}}  \shline
&Vanilla&KD \cite{Hinton2015DistillingTK}&AT \cite{Zagoruyko2017PayingMA}&RKD \cite{Park2019RelationalKD}& SCKD$^*$ \cite{Chen2020CrossLayerDW}&CRD \cite{Tian2020ContrastiveRD}&CRD+KD& SAD$^*$ \cite{Ji2021ShowAA}&CC$^*$ \cite{Peng2019CorrelationCF}& ICKD-C (Ours)& Teacher\\ \hline
w/ $\mathcal{L}_{\rm KL}$  &      &$\checkmark$      &      &$\checkmark$       &$\checkmark$       & &$\checkmark$  &     &$\checkmark$      & $\checkmark$  &  \\ 
Top-1                   &70.04  &70.68  &70.59  &71.34  &70.87  &71.17  &71.38 &71.38  &70.74  & \textbf{72.19} & 73.31\\
Top-5                   &89.48  &90.16  &89.73  &90.37  &NA     &90.13  &90.49 &NA     &NA &\textbf{90.72} & 91.42\\ \hline

\end{tabular}
}\label{imagenet}
\vspace{-3mm}
\end{table*}


\section{Experiments}
We evaluate the effectiveness of the proposed model on two vision tasks: image classification and semantic segmentation. For image classification, we conduct the experiments on Cifar-100 and ImageNet. To verify the generalization of our framework, we further conduct experiments of semantic segmentation on the large-scale benchmark Pascal VOC.
\subsection{Datasets}
\textit{ImageNet.} This dataset has about 1.2\textbf{M} training samples labeled into 1,000 categories. The images are resized to 224$\times$224\ for both training and testing.
Usually, the performance of a model is measured by Top-1 and Top-5 classification accuracy. 

\textit{Cifar-100.} This dataset contains 50,000 training images and 10,000 testing images, labeled into 100 categories. Each image is of size $32\times32\times3$. Top-1 classification accuracy is adopted to measure the model.

\textit{Pascal VOC.} This dataset contains 20 foreground object classes plus an extra background class. It has 1,464 images for training, 1,499 images for validation and 1,456 images for testing. We also include the coarse annotated training images from \cite{Hariharan2011SemanticCF}, resulting in 10,582 training images in total. We employ mean Intersection over Union (mIoU) to evaluate the effectiveness of the proposed model.

\subsection{Implementation Details}
For image classification, the feature map before the global average pooling layer is used for distillation. We empirically set the weight factors of $\beta_1$ and $\beta_2$ in Eq. \ref{eq:objective} to 1 and 2.5, respectively. On Cifar-100, the SGD optimizer \cite{sutskever2013importance} is applied to train the student model with Nesterov momentum and a batch size of 64. The initial learning rate is 5e-2 and decayed by 0.1 at epochs 150, 180, and 210 with 240 epochs in total. In terms of ImageNet, we use the AdamW optimizer \cite{loshchilov2017decoupled} to train the network for 100 epochs with a total batch size of 256. The initial learning rate is 2e-4 reduced by 0.1 at epochs 30, 60, and 90.

As to semantic segmentation, we distill the knowledge on the last BN \cite{ioffe2015batch} layer of DeeplabV3+, whose feature map size is 256$\times$129$\times$129. The weight $\alpha$ in Eq. \ref{eq:seg_objective} is set to 20. All students are trained for 100 epochs with a batch size of 12. We use the SGD optimizer with an initial learning rate of 0.007. And the learning rate decays according to the cosine annealing scheduler.
\subsection{Image Classification}
\noindent\textbf{Results on Cifar-100.} 
We evaluate the proposed method in a variety of network architectures, including VGG \cite{simonyan2014very}, ResNet \cite{He2016DeepRL} and its variants \cite{Zagoruyko2016WideRN}. As shown in Table \ref{tab:cifar_sota}, our method outperforms other methods by a large margin. 
In the setting of distillation from WRN-40-2 to WRN-16-2, we achieve 75.57\% Top-1 accuracy which is close to the teacher's performance 75.61\%. 
We also compare to the methods that are more relevant to us, including layer-wise relation \cite{Passalis2018LearningDR,Tung2019SimilarityPreservingKD,Peng2019CorrelationCF,Park2019RelationalKD} and those based on Gram matrix \cite{Yim2017AGF,huang2017like} which measures the relation between pixel-wise positions. In terms of layer-wise relational knowledge distillation methods, we outperform all the state-of-the-arts consistently. For example, in the distillation setting from ResNet56 to ResNet20, our method achieves an accuracy of 71.69\% which greatly exceeds the second best method. This consistency proves the important role of feature diversity in knowledge distillation. As can be observed, the other state-of-the-arts ranked inconsistently and the traditional KD \cite{Hinton2015DistillingTK} ranked the second place at most time. We can say that the contribution of mining the relationship of the layer-wise features is less than the guarantee of feature diversity. 

We further explore the potential of inter-channel correlation by distillation across different network architectures, including MobileNetV2 \cite{Sandler2018MobileNetV2IR}, ShuffleNetV1 \cite{zhang2018shufflenet},  and ShuffleNetV2 \cite{Sandler2018MobileNetV2IR} (See Fig. \ref{fig:diff_arch}). 
The characteristic of an ideal knowledge distillation method is that it can transfer the general knowledge regardless of the specific architecture. We find that some methods even deteriorated the performance of the students. Cho \etal \cite{Cho2019OnTE} has pointed out that the students may fail to catch up with the teachers if their learning capacities mismatch. In the case that VGG13 is adopted as the teacher of MobileNetV2, many methods fail to improve the performance of the student. The situation even became worse when trying unilaterally to lead the student to learn the high-response area of the teacher's features given that AT \cite{Zagoruyko2017PayingMA} dropped 5\% compared to the vanilla student. On the contrary, due to the characteristic of being invariant to the spatial dimension, ICKD-C can be adopted to transfer knowledge across different architectures, and its performance is always better than other methods. For instance, the transferred layer of VGG13 has a different feature map size from that of MobileNetV2, our method surpasses the second-best about 1\% accuracy. 

\begin{figure*}[t]
\begin{center}
\includegraphics[width=.98\linewidth]{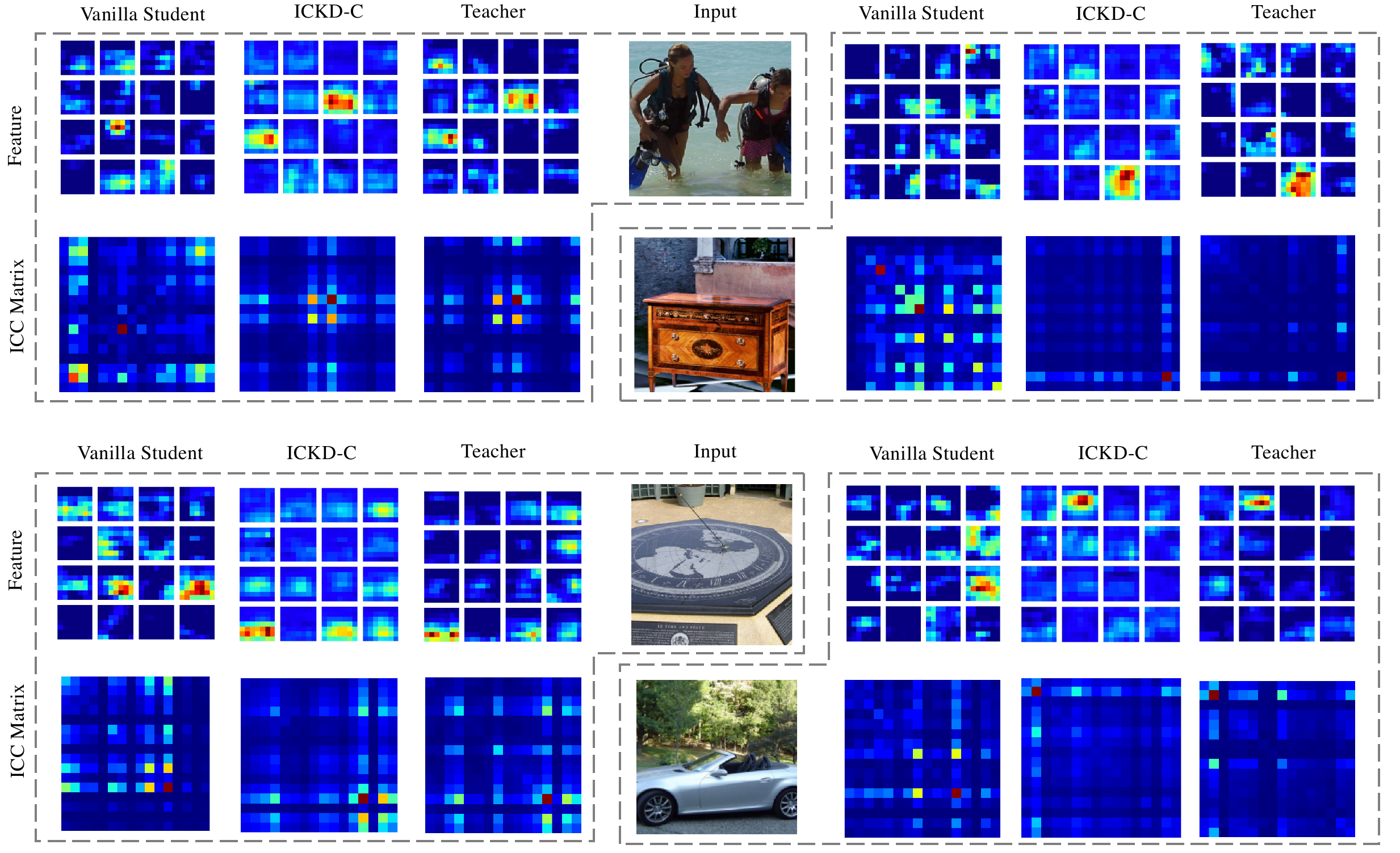}
\end{center}
\vspace{-6mm}
   \caption{\textbf{Visualization of the features and the ICC matrices}. We have visualized the feature maps and the corresponding ICC matrices of the vanilla student, our model (ICKD-C) and the teacher, respectively. The four input images are sampled from ImageNet testing set. The teacher architecture is ResNet34 and the student architecture is ResNet18. Without loss of generality, we orderly select 16 feature maps extracted from the 4-th block (\ie, the distillation layer) of the network. The results show that our model possesses the similar feature diversity and pattern with the teacher, demonstrating that learning inter-channel correlation can effectively preserve feature diversity.}
\label{fig:iccmatrix}
\vspace{-4mm}
\end{figure*}

\noindent\textbf{Results on ImageNet.}
We evaluate our method on the larger scale dataset ImageNet \cite{deng2009imagenet}. Note that \cite{Park2019RelationalKD} additionally applied the rotation, horizontal flipping, and color jittering for data augmentation.
To compare with other works more fairly, we choose ResNet34 as the teacher network and ResNet18 as the student network.
The result is presented in Table \ref{imagenet}. Again, our method consistently outperforms all methods by a significant margin. Our result is remarkable in that it achieves an accuracy rate of more than 72\% in the existing literature for the first time.

We visualize the ICC matrices of the student network and teacher network (See Fig. \ref{fig:iccmatrix}). At first, the feature maps of student and teacher show great differences both in the inter-channel correlation and the response on a single channel. However, after distillation, they have become similar in addition to the ICC matrix, and the response on a single channel is also closer. 
According to the visualization of the feature channels, we can say the student can effectively preserve the feature diversity and has a similar feature pattern with the teacher. More results are displayed in Appendix.

\subsection{Semantic Segmentation}
 Semantic segmentation is a promising but computation-consuming application. 
 Yet methods based on knowledge distillation are rarely successfully applied to semantic segmentation. In this section, we present the experiments on the Pascal VOC semantic segmentation in the setting of knowledge distillation. Specifically, we deploy the ResNet101 as teacher backbone and transfer to student backbones ResNet18 and MoobileNetV2. The DeepLabV3+ \cite{chen2018encoder} is chosen as the baseline model. Semantic segmentation aims at pixel-level classification, which is more challenging than image classification. The result is displayed in Table \ref{tab:seg}. It shows that we can prompt the student by a large margin (from 72.07\% to 75.01\%), which demonstrates that our method can learn rich representation for different downstream tasks. Particularly, our method bridges the gap between the cumbersome teacher and the inferior student, making it feasible to deploy segmentation models on edge devices.

\vspace{-2mm}
\subsection{The potential of a better
teacher}\label{sec:ablation}
\begin{figure} [h!]
    \centering 
    \includegraphics[width=.9\linewidth]{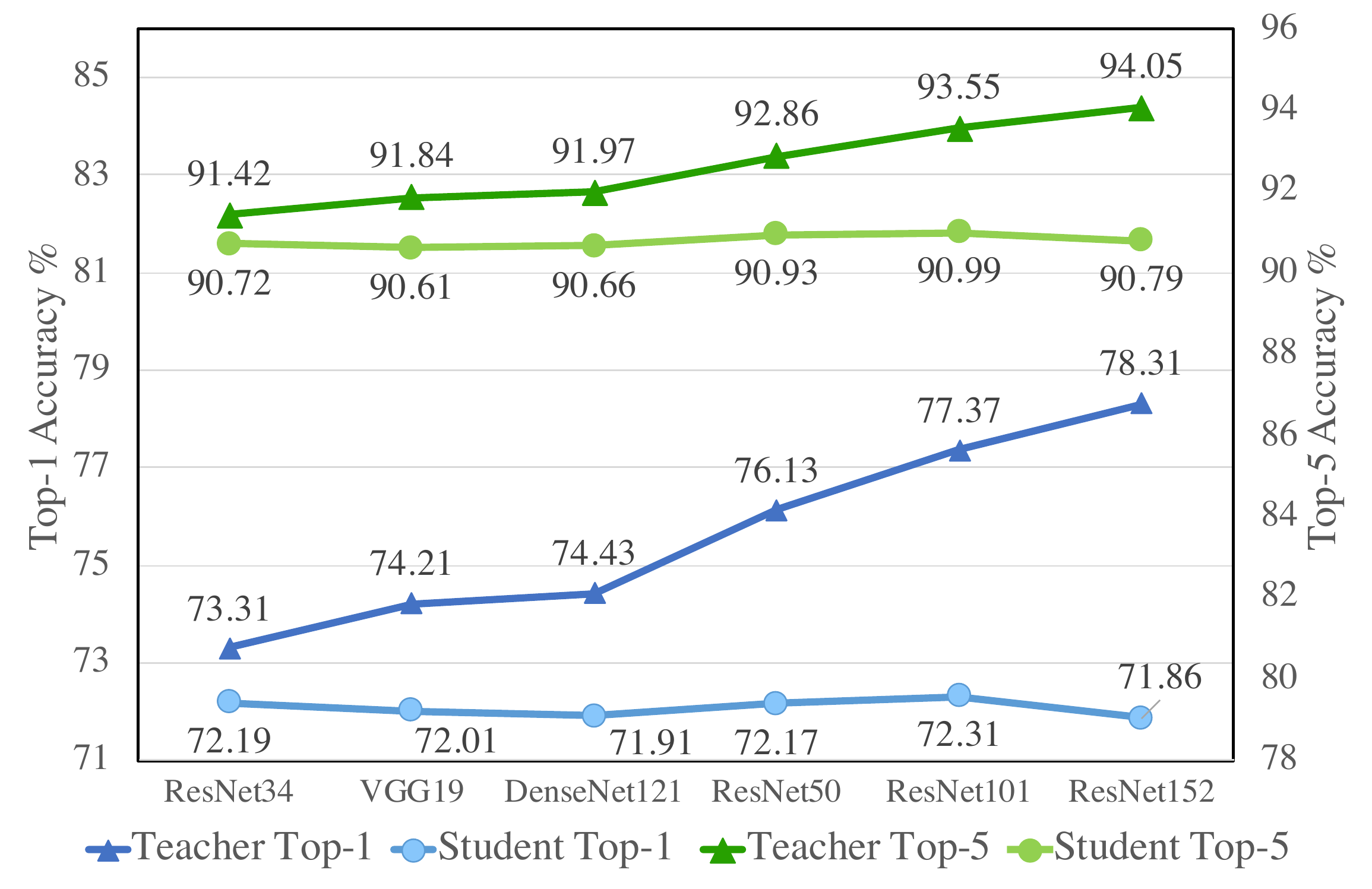}\vspace{-3mm}
    \caption{Accuracy (\%) of the same student (ResNet18) guided by different teachers on ImageNet.} 
    \label{fig:diff_teacher} 
    \vspace{-0mm}
\end{figure}

\vspace{-1mm}
An assumption is that the better the teachers are, the better the students we would have. This assumption seems plausible yet has been demonstrated unpractical because the students may not be able to catch up with the teachers \cite{Cho2019OnTE}. We use several teacher networks individually to train the same student network (ResNet18) to see the possible improvements. As shown in Fig. \ref{fig:diff_teacher}, although all of the teachers can bring considerable performance gain to the student, the heavier teachers could not consistently prompt the Top-1 accuracy than the lightweight one. The student can achieve the best performance (Top-1 72.31\%) when ResNet101 is used as the teacher and the second-best performance (Top-1 72.19\%) when ResNet34 is adopted. Except for ResNet101, teachers better than ResNet34 could not bring further improvement. Interestingly, the best teacher (ResNet152) couldn't obtain a considerable student model compared with others, which may be caused by the huge difference between their channel numbers (2048 for ResNet152 and 512 for ResNet18). We may say that it is unnecessary to employ a very cumbersome teacher network for knowledge distillation since it cannot bring further improvement consistently and spends more cost on pre-training.
\begin{table}[h]
    \centering
    \caption{Performance on semantic segmentation in terms of mIoU (\%) on the validation set of Pascal VOC.}\vspace{-3mm}
    \resizebox{0.9\columnwidth}{!}{\tablestyle{10pt}{1.05}
    \begin{tabular}{@{}l|cc@{}} \shline
         Model&ResNet18&MobileNetV2 \\  \hline
         Vanilla&72.07&68.46 \\ 
         KD \cite{Hinton2015DistillingTK}& 73.74 & 71.73\\ 
         FitNet \cite{Romero2015FitNetsHF}& 73.31 & 69.23 \\ 
         AT \cite{Zagoruyko2017PayingMA}& 73.01 & 71.39 \\ 
         Overhaul \cite{Heo2019ACO}& 73.98 & 71.19 \\
         ICKD-S (Ours) & \textbf{75.01} & \textbf{72.79} \\ \hline
    \end{tabular}
    \label{tab:seg}
    }
    \vspace{-6mm}
\end{table}
\subsection{Ablation Study}\label{sec:ablation}
\vspace{-1mm}
Firstly, we study the impact of the linear transformation layer $C_l$ on Cifar-100. Intuitively, the linear transformation module may hinder the process of inter-channel correlation knowledge distillation. However, the results presented in Table \ref{tab:cifar_ablation} show that the 1$\times$1 linear transformation leads to a minor gain in most cases. This phenomenon is also observed by Wang \etal \cite{wang2019distilling} \cready{in which} linear transformation acts as an adaptor between the teacher and student.

Secondly, we study the impact of the weight factor $\beta_2$ in Eq. \ref{eq:objective}. 
In order to exclude the influence of $\mathcal{L}_{\rm KL}$ ( Eq. \ref{eq:kl}) and verify the effectiveness of $\mathcal{L}_{\rm CC}$ separately, we set $\beta_1$ to zero. The results in Table \ref{tab:beta2} illustrate that our method is still impressive without joining $\mathcal{L}_{\rm KL}$ on ImageNet (71.59\% Top-1 accuracy, which also surpasses the methods in Table \ref{imagenet}). And it is also very robust to $\beta_2$. \cready{We perform the ICC matrix transferring (ResNet34{$\rightarrow$}ResNet18) at different stages. The stage numbers are indicated by the subscript. When a single layer is used, our strategy is the best. When multiple stages are get involved with training, $\rm{S_{3,4}}$ achieves the best Top-1 accuracy (See Table \ref{tab:diff_layer}). In addition, we also conduct the experiments under different loss functions and kernel functions (See Table \ref{tab:diff_func}).}

Lastly, the grid-level inter-channel correlation proposed in Sec. \ref{sec:seg} is able to bring more performance improvements. Recall that we divide the whole feature map into $n\times m$ parts and if it is set to 1$\times$1, this variant degrades to the ICKD-C without $\mathcal{L}_{\rm KL}$. We conduct some experiments under different settings of $n\times m$ to see its effect. As can be observed in Table \ref{tab:seg_nxn}, our proposed ICKD-C without $\mathcal{L}_{\rm KL}$ still improve the student (ResNet18) by 2.07\% (from 72.07 to 74.14) and it can further boost the student to 75.01 after dividing the feature map into $32 \times 32$ patches. Generally, meshing the feature map can consistently improve the performance, but it is not the finer the better. Besides, the finer grid means more training cost. Table \ref{tab:seg_nxn_time} illustrated the GPU hours cost of training the segmentation model 100 epochs with 2$\times$NVIDIA 2080Ti.
\begin{table}[h!]
    \caption{ICC Transferring at different places on ImageNet.}
    \vspace{-2mm}
    \centering
\resizebox{1.0\columnwidth}{!}{
    \begin{tabular}{c|cccc|cccc}
    \shline
         &$\rm{S_1}$&$\rm{S_2}$&$\rm{S_3}$&$\rm{S_4}$ (ours)&$\rm{S_{1,4}}$&$\rm{S_{2,4}}$&$\rm{S_{3,4}}$&$\rm{S_{1,2,3,4}}$  \\
         Top-1&70.49&70.50&70.87&\textbf{72.16}&72.31&72.20&72.33&72.26 \\
         Top-5&89.47&89.53&89.59&\textbf{90.75}&90.67&90.71&90.55&90.64\\
    \hline
    \end{tabular}}
    \label{tab:diff_layer}
\end{table}
\vspace{-5mm}
\begin{table}[h!]
    \caption{Different loss functions and kernel functions.}
    \vspace{-2mm}
    \centering
\resizebox{1.0\columnwidth}{!}{
    \begin{tabular}{c|cc|cc} 
    \shline
         &\multicolumn{2}{c|}{Loss Functions} &\multicolumn{2}{c}{Kernel Functions}\\
         &L2 (ours)&Smooth L1&Gaussian kernel&Polynomial kernel  \\
         Top-1& 72.16&72.29&70.63&72.25 \\
         Top-5& 90.75&90.75&89.73&90.80 \\
    \hline
    \end{tabular}}
    \label{tab:diff_func}
\end{table}
\begin{table}[!h]
\centering
\caption{Ablation on Cifar-100.}
\vspace{-2mm}
\resizebox{0.95\columnwidth}{!}{\tablestyle{10pt}{1.05}
\begin{tabular}{@{}ll|cc@{}} \shline
Teacher&Student&w/o Linear&w/ Linear \\\hline
 WRN-40-2 & WRN-16-2 &75.10 &\textbf{75.64}    \\
 WRN-40-2  & WRN-40-1 &73.87 &\textbf{74.33} \\
 ResNet56 & ResNet20 &71.72&\textbf{71.76} \\
 ResNet110 & ResNet20 &70.96&\textbf{71.68}\\
 ResNet110  & ResNet32&\textbf{73.90}&73.89 \\
ResNet32$\times$4 & ResNet8$\times$4 &74.40 &\textbf{75.25} \\
 VGG13 & VGG8 &\textbf{73.85}  &73.42 \\ \hline
\end{tabular}
}
\label{tab:cifar_ablation}
\end{table}
\vspace{-3mm}
\begin{table}[!h]
    \centering
    \caption{Top-1 accuracy(\%) under different $\beta_2$ on ImageNet.}
    \vspace{-2mm}
    \resizebox{1.0\columnwidth}{!}{\tablestyle{14pt}{1.05}
    \begin{tabular}{@{}c|cccc@{}}  \shline
         $\beta_2$&$0.2$&$1.0$&$2.0$&$4.0$\\ \hline
         ACC.&71.09&71.17&\textbf{71.59}&71.34\\ \hline
    \end{tabular}
    \label{tab:beta2}
    }    
\end{table}
\begin{table}[h]
    \centering
    \caption{mIoU(\%) under different settings of $n \times m$ on Pascal VOC.}
    \vspace{-2mm}
    \resizebox{1\columnwidth}{!}{\tablestyle{10pt}{1.05}
    \begin{tabular}{@{}c|cccc@{}}  \shline
         {$n\times m$}&$1\times1$&$4\times4$&$16\times16$&$32\times32$ \\ \hline
         ResNet18&74.14&74.97&74.74&\textbf{75.01}\\ 
         MobileNetV2&72.10&72.26&\textbf{72.79}&72.58\\ \hline
    \end{tabular}
    \label{tab:seg_nxn}
    }
    \vspace{-0mm}
\end{table}

\begin{table}[h]
    \centering
    \caption{Training cost (GPU Hours) under different settings of $n \times m$ on Pascal VOC.}
    \vspace{-2mm}
    \resizebox{1\columnwidth}{!}{\tablestyle{10pt}{1.05}
    \begin{tabular}{@{}c|cccc@{}}  \shline
         {$n\times m$}&$1\times1$&$4\times4$&$16\times16$&$32\times32$ \\ \hline
         ResNet18&25.8&25.9&31.8&157.4\\ 
         MobileNetV2&24.5&24.7&30.4&155.9\\ \hline
    \end{tabular}
    \label{tab:seg_nxn_time}
    }
    \vspace{-4mm}
\end{table}

\section{Conclusion}
This work presents a method for knowledge distillation that explores the inter-channel correlation to mimic the feature diversity of the teacher network. In addition to image classification, we introduce the grid-level inter-channel correlation for semantic segmentation that most prior works do not pay attention to. We empirically demonstrate the effectiveness of the proposed method on a variety of network architectures and achieve the state-of-the-art in two vision tasks (image classification and semantic segmentation). 
Besides, the computation of the proposed ICC matrix is invariant to feature spatial dimensions and able to distill generic knowledge across different network architectures.
\section*{Acknowledgement}
This work was supported by the funding of ``Leading Innovation Team of the Zhejiang Province" (2018R01017) and Australian Research Council (ARC) Discovery Early Career Researcher Award (DECRA) under DE190100626.

\newpage
{\small
\bibliographystyle{ieee_fullname}
\bibliography{egbib}

\begin{thebibliography}{10}\itemsep=-1pt

\bibitem{bucilua2006model}
Cristian Bucilua, Rich Caruana, and Alexandru Niculescu-Mizil.
\newblock Model compression.
\newblock {\em Proceedings of the 12th ACM SIGKDD international conference on
  Knowledge discovery and data mining}, pages 535--541, 2006.

\bibitem{Chen2020CrossLayerDW}
Defang Chen, Jian-Ping Mei, Yeliang Zhang, Can Wang, Zhe Wang, Yan Feng, and
  Chun Chen.
\newblock Cross-layer distillation with semantic calibration.
\newblock {\em ArXiv}, abs/2012.03236, 2020.

\bibitem{chen2018encoder}
Liang-Chieh Chen, Yukun Zhu, George Papandreou, Florian Schroff, and Hartwig
  Adam.
\newblock Encoder-decoder with atrous separable convolution for semantic image
  segmentation.
\newblock In {\em Proceedings of the European conference on computer vision
  (ECCV)}, pages 801--818, 2018.

\bibitem{Cho2019OnTE}
J.~H. Cho and B. Hariharan.
\newblock On the efficacy of knowledge distillation.
\newblock {\em 2019 IEEE/CVF International Conference on Computer Vision
  (ICCV)}, pages 4793--4801, 2019.

\bibitem{deng2009imagenet}
Jia Deng, Wei Dong, Richard Socher, Li-Jia Li, Kai Li, and Li Fei-Fei.
\newblock Imagenet: A large-scale hierarchical image database.
\newblock In {\em 2009 IEEE conference on computer vision and pattern
  recognition}, pages 248--255. Ieee, 2009.

\bibitem{Gatys2016ImageST}
Leon~A. Gatys, Alexander~S. Ecker, and M. Bethge.
\newblock Image style transfer using convolutional neural networks.
\newblock {\em 2016 IEEE Conference on Computer Vision and Pattern Recognition
  (CVPR)}, pages 2414--2423, 2016.

\bibitem{Han2020GhostNetMF}
Kai Han, Yunhe Wang, Q. Tian, Jianyuan Guo, Chunjing Xu, and C. Xu.
\newblock Ghostnet: More features from cheap operations.
\newblock {\em 2020 IEEE/CVF Conference on Computer Vision and Pattern
  Recognition (CVPR)}, pages 1577--1586, 2020.

\bibitem{Hariharan2011SemanticCF}
Bharath Hariharan, Pablo Arbel{\'a}ez, Lubomir~D. Bourdev, Subhransu Maji, and
  Jitendra Malik.
\newblock Semantic contours from inverse detectors.
\newblock {\em 2011 International Conference on Computer Vision}, pages
  991--998, 2011.

\bibitem{He2016DeepRL}
Kaiming He, X. Zhang, Shaoqing Ren, and Jian Sun.
\newblock Deep residual learning for image recognition.
\newblock {\em 2016 IEEE Conference on Computer Vision and Pattern Recognition
  (CVPR)}, pages 770--778, 2016.

\bibitem{he2019knowledge}
Tong He, Chunhua Shen, Zhi Tian, Dong Gong, Changming Sun, and Youliang Yan.
\newblock Knowledge adaptation for efficient semantic segmentation.
\newblock In {\em Proceedings of the IEEE/CVF Conference on Computer Vision and
  Pattern Recognition}, pages 578--587, 2019.

\bibitem{Heo2019ACO}
Byeongho Heo, Jeesoo Kim, Sangdoo Yun, H. Park, N. Kwak, and J. Choi.
\newblock A comprehensive overhaul of feature distillation.
\newblock {\em 2019 IEEE/CVF International Conference on Computer Vision
  (ICCV)}, pages 1921--1930, 2019.

\bibitem{Hinton2015DistillingTK}
Geoffrey~E. Hinton, Oriol Vinyals, and J. Dean.
\newblock Distilling the knowledge in a neural network.
\newblock {\em ArXiv}, abs/1503.02531, 2015.

\bibitem{huang2017like}
Zehao Huang and Naiyan Wang.
\newblock Like what you like: Knowledge distill via neuron selectivity
  transfer.
\newblock {\em arXiv preprint arXiv:1707.01219}, 2017.

\bibitem{ioffe2015batch}
Sergey Ioffe and Christian Szegedy.
\newblock Batch normalization: Accelerating deep network training by reducing
  internal covariate shift.
\newblock In {\em International conference on machine learning}, pages
  448--456. PMLR, 2015.

\bibitem{Ji2021ShowAA}
Mingi Ji, Byeongho Heo, and S. Park.
\newblock Show, attend and distill: Knowledge distillation via attention-based
  feature matching.
\newblock {\em ArXiv}, abs/2102.02973, 2021.

\bibitem{lee2018self}
Seung~Hyun Lee, Dae~Ha Kim, and Byung~Cheol Song.
\newblock Self-supervised knowledge distillation using singular value
  decomposition.
\newblock In {\em Proceedings of the European Conference on Computer Vision
  (ECCV)}, pages 335--350, 2018.

\bibitem{liu2019structured}
Yifan Liu, Ke Chen, Chris Liu, Zengchang Qin, Zhenbo Luo, and Jingdong Wang.
\newblock Structured knowledge distillation for semantic segmentation.
\newblock In {\em Proceedings of the IEEE/CVF Conference on Computer Vision and
  Pattern Recognition}, pages 2604--2613, 2019.

\bibitem{loshchilov2017decoupled}
Ilya Loshchilov and Frank Hutter.
\newblock Decoupled weight decay regularization.
\newblock {\em arXiv preprint arXiv:1711.05101}, 2017.

\bibitem{Park2019RelationalKD}
Wonpyo Park, Dongju Kim, Yan Lu, and Minsu Cho.
\newblock Relational knowledge distillation.
\newblock {\em 2019 IEEE/CVF Conference on Computer Vision and Pattern
  Recognition (CVPR)}, pages 3962--3971, 2019.

\bibitem{Passalis2018LearningDR}
N. Passalis and A. Tefas.
\newblock Learning deep representations with probabilistic knowledge transfer.
\newblock In {\em ECCV}, 2018.

\bibitem{Peng2019CorrelationCF}
Baoyun Peng, Xiao Jin, Jiaheng Liu, Shunfeng Zhou, Y. Wu, Y. Liu, Dong sheng
  Li, and Z. Zhang.
\newblock Correlation congruence for knowledge distillation.
\newblock {\em 2019 IEEE/CVF International Conference on Computer Vision
  (ICCV)}, pages 5006--5015, 2019.

\bibitem{Romero2015FitNetsHF}
A. Romero, Nicolas Ballas, S. Kahou, Antoine Chassang, C. Gatta, and Yoshua
  Bengio.
\newblock Fitnets: Hints for thin deep nets.
\newblock {\em CoRR}, abs/1412.6550, 2015.

\bibitem{Sandler2018MobileNetV2IR}
Mark Sandler, A. Howard, Menglong Zhu, A. Zhmoginov, and Liang-Chieh Chen.
\newblock Mobilenetv2: Inverted residuals and linear bottlenecks.
\newblock {\em 2018 IEEE/CVF Conference on Computer Vision and Pattern
  Recognition}, pages 4510--4520, 2018.

\bibitem{simonyan2014very}
Karen Simonyan and Andrew Zisserman.
\newblock Very deep convolutional networks for large-scale image recognition.
\newblock {\em arXiv preprint arXiv:1409.1556}, 2014.

\bibitem{Sun2019PatientKD}
S. Sun, Yu Cheng, Zhe Gan, and Jingjing Liu.
\newblock Patient knowledge distillation for bert model compression.
\newblock In {\em EMNLP/IJCNLP}, 2019.

\bibitem{sutskever2013importance}
Ilya Sutskever, James Martens, George Dahl, and Geoffrey Hinton.
\newblock On the importance of initialization and momentum in deep learning.
\newblock In {\em International conference on machine learning}, pages
  1139--1147. PMLR, 2013.

\bibitem{Tian2020ContrastiveRD}
Yonglong Tian, Dilip Krishnan, and Phillip Isola.
\newblock Contrastive representation distillation.
\newblock {\em ICLR}, 2020.

\bibitem{Tung2019SimilarityPreservingKD}
F. Tung and G. Mori.
\newblock Similarity-preserving knowledge distillation.
\newblock {\em 2019 IEEE/CVF International Conference on Computer Vision
  (ICCV)}, pages 1365--1374, 2019.

\bibitem{wang2019distilling}
Tao Wang, Li Yuan, Xiaopeng Zhang, and Jiashi Feng.
\newblock Distilling object detectors with fine-grained feature imitation.
\newblock In {\em Proceedings of the IEEE/CVF Conference on Computer Vision and
  Pattern Recognition}, pages 4933--4942, 2019.

\bibitem{Wang2020IntraclassFV}
Yukang Wang, W. Zhou, T. Jiang, X. Bai, and Yongchao Xu.
\newblock Intra-class feature variation distillation for semantic segmentation.
\newblock In {\em ECCV}, 2020.

\bibitem{wei2019building}
Zhen Wei, Jingyi Zhang, Li Liu, Fan Zhu, Fumin Shen, Yi Zhou, Si Liu, Yao Sun,
  and Ling Shao.
\newblock Building detail-sensitive semantic segmentation networks with
  polynomial pooling.
\newblock In {\em Proceedings of the IEEE/CVF Conference on Computer Vision and
  Pattern Recognition}, pages 7115--7123, 2019.

\bibitem{Yim2017AGF}
Junho Yim, Donggyu Joo, Jihoon Bae, and Junmo Kim.
\newblock A gift from knowledge distillation: Fast optimization, network
  minimization and transfer learning.
\newblock {\em 2017 IEEE Conference on Computer Vision and Pattern Recognition
  (CVPR)}, pages 7130--7138, 2017.

\bibitem{Zagoruyko2016WideRN}
Sergey Zagoruyko and Nikos Komodakis.
\newblock Wide residual networks.
\newblock {\em ArXiv}, abs/1605.07146, 2016.

\bibitem{Zagoruyko2017PayingMA}
Sergey Zagoruyko and Nikos Komodakis.
\newblock Paying more attention to attention: Improving the performance of
  convolutional neural networks via attention transfer.
\newblock {\em ArXiv}, abs/1612.03928, 2017.

\bibitem{zhang2018shufflenet}
Xiangyu Zhang, Xinyu Zhou, Mengxiao Lin, and Jian Sun.
\newblock Shufflenet: An extremely efficient convolutional neural network for
  mobile devices.
\newblock In {\em Proceedings of the IEEE conference on computer vision and
  pattern recognition}, pages 6848--6856, 2018.

\end{thebibliography}
}

\end{document}